# A Data-Driven Analytical Framework of Estimating Multimodal Travel Demand Patterns using Mobile Device Location Data

Chenfeng Xiong[1], Aref Darzi[2], Yixuan Pan[2], Sepehr Ghader[3], Lei Zhang[4]*

1. Assistant Research Professor, Department of Civil and Environmental Engineering, University of Maryland, College Park, MD, 20742.
2. Graduate Research Assistant, Department of Civil and Environmental Engineering, University of Maryland, College Park, MD, 20742, USA
3. Research Scientist, Department of Civil and Environmental Engineering, University of Maryland, College Park, MD, 20742, USA
4. (Corresponding Author) Herbert Rabin Distinguished Professor, Department of Civil and Environmental Engineering, University of Maryland, College Park, MD, 20742, USA
(Correspondence) Phone: (301)405-2881; Email: lei@umd.edu

**ABSTRACT**

While benefiting people's daily life in so many ways, smartphones and their location-based services are generating massive mobile device location data that has great potential to help us understand travel demand patterns and make transportation planning for the future. While recent studies have analyzed human travel behavior using such new data sources, limited research has been done to extract multimodal travel demand patterns out of them. This paper presents a data-driven analytical framework to bridge the gap. To be able to successfully detect travel modes using the passively collected location information, we conduct a smartphone-based GPS survey to collect ground truth observations. Then a jointly trained single-layer model and deep neural network for travel mode imputation is developed. Being "wide" and "deep" at the same time, this model combines the advantages of both types of models. The framework also incorporates the multimodal transportation network in order to evaluate the closeness of trip routes to the nearby rail, metro, highway and bus lines and therefore enhance the imputation accuracy. To showcase the applications of the introduced framework in answering real-world planning needs, a separate mobile device location data is processed through trip end identification and attribute generation, in a way that the travel mode imputation can be directly applied. The estimated multimodal travel demand patterns are then validated against typical household travel surveys in the same Washington D.C. and Baltimore Metropolitan Regions.

# 1. BACKGROUND

Thanks to the rapidly evolving smartphone industry and mobile computing technology, mobile device location data has never been so readily available before. According to the Pew Research Center, the United States has around 223 million smartphone users in 2017 ([Mobile Fact Sheet](#)). More than three-quarters of Americans (77%) now own a smartphone, with lower-income Americans and senior citizens above the age of 50 exhibiting a sharp uptick in ownership over the past years. These devices are generating a massive amount of location data continuously through the widespread use of location-based service (LBS) via Wi-Fi hotspots, cellular towers, Global Positioning System (GPS)-based technologies, and GPS-enabled applications on these smartphone devices. This ubiquitous LBS data provides an opportunity to innovatively and accurately observe individuals' travel behavior and model the overall travel demand patterns for a region, a state, and even an entire country.

Existing methodologies of estimating travel demand, such as the widely used four-step method and the activity-based method, have to rely heavily on statistical sampling of empirical evidence, e.g. a regional household travel survey (see, for example, Daganzo, 1980; Richardson et al., 1995; Pendyala et al., 2005; Stopher and Greaves, 2007). These survey-powered data sources have certain limitations. They are costly to collect and thus may only be sampled once every 5~10 years with low sampling rate. In comparison, location-based service data represents a new and alternative big-data resource to capture personal-level movements dynamically and continuously on relatively larger samples. This new opportunity to supplement and enhance the current transportation and urban planning processes is certainly worth research attention.

Starting from the seminal works using call detail records to identify human mobility patterns (Gonzalez et al., 2008; Montjoye et al., 2013; Calabrese, et al., 2013; Iqbal, et al., 2014), researchers have found that passively collected data from mobile devices can successfully reflect individuals' travel patterns and can be used to predict travel demand. Schneider et al. (2013) found that up to 90% of the population in surveys and mobile device datasets can be represented by a handful of unique networks called motifs. The mobile device location data has also been applied in the contexts of resilience and evacuation (Lu, et al. 2012) and national pandemics and disease control (Belik, et al., 2011; Wesolowski et al., 2012). Toole et al. (2015) made an attempt to use phone call detail records to build origin-destination matrices and successfully demonstrated the feasibility in several metropolitan areas internationally. One of the latest attempts made by Bachir et al., (2019) inferred multimodal origin-destination flows using call records. Among these studies, a common limitation lies in the accurate detection of travel modes for estimating demand patterns. Due to the nature of the passively collected mobile device data, there is no ground-truth mode label to guide the imputation of travel modes. For the transportation planning of an urban area with multimodal transportation networks, this is certainly an important research gap that needs to be addressed.

We are thus motivated to develop a data-driven framework to enable the multimodal travel demand estimation in order to make the mobile device location data even more useful. To achieve this goal, we must integrate a smartphone GPS survey, new machine learning algorithms, and available transportation and urban-level data sources in the domain to better understand the multimodal travel behavior and trajectories. At the same time, we need to keep the model attributes simple in order to be able to apply the model on big data location observations. The supplementary information, ground truth observations, and new models ensure the successful detection of travel modes for the individual trips extracted passively and anonymously from the big data.

Travel behavioral analysis and mode detection based on GPS raw data drew increasing research attention in the past decade. Researchers have explored artificial intelligence (AI) methods to cope with passively collected location-based data, including hierarchical Markov models (Liao et al., 2007), Decision Tree (e.g. Stenneth 2011; Zheng, 2008), Neural Network (Gonzalez et al., 2010; Byon, 2014; Yang, 2015; Dabiri and Heaslip, 2018), Naïve Bayes and Bayesian Networks (Xiao, et al., 2015), Support Vector Machines (e.g. Zhang, 2011), Random Forest (e.g. Witayangkurn, et al., 2013; Lari, 2015), ensemble models (Hasnat and Hasan, 2018), etc. Wu et al. (2016), Chen et al., (2016), and Huang et al., (2019) have conducted thorough syntheses of existing studies on this topic. Data, methodologies, and research outcomes are reviewed and compared. Overall, the current practices can detect car mode with high accuracy (with high-resolution GPS traces which also draws battery concerns). However, the detection of bus/metro/subway modes is not as satisfying. One possible methodological limitation that could lead to this is the single-layer AI representation. The single-layer neurons or rules often cannot make enough generalization on a high-dimensional problem. To generalize to unobserved feature combinations for bus or metro modes, a multi-layer deep neural network (DNN) can be used, which perform efficiently with much fewer nodes in each layer.

To effectively detect the travel mode for each trip, feature construction for classification is critical in providing useful information, and preferably travel mode-specific knowledge, to improve the detection accuracy. In addition to the traditional features used in the literature (e.g. average speed, maximum speed, trip distance, etc.), this study constructed two innovative features based on network data: the distance to the closest rail line (both underground and aboveground) and the distance to the closest bus line. Stenneth et al. (2011) and Witayangkurn et al., (2013) first proposed to use transportation network information in travel mode detection. However, their papers did not consider metro (underground) detection and bus line information. In our smartphone GPS survey in the Washington-Baltimore's Metro systems, GPS locations can be recorded at metro lines and stations, which makes underground metro detection possible. To the best knowledge of the authors, this paper is one of the first to use a comprehensive set of multimodal network data to infer highway, Metro, and bus modes.

There are two major contributions of this paper. First of all, we present a study employing a jointly trained single-layer model and deep neural network for travel mode detection. This model combines the advantages of both types of models and is shown to greatly improve the performance. To further improve the accuracy, the paper adopts multimodal network measurements and assembled all available GTFS bus data from 31 regional and local agencies and bus service providers. These features are found extremely useful to detect bus trips and Metro trips that contain underground portions and/or have signal quality issues. These collectively contribute to a model with high detection accuracy at the individual trip level. With the developed and empirically-trained wide-and-deep learning model playing a central role in the proposed data-driven analytical framework, the paper fills a critical data gap of lacking travel mode information in the emerging mobile device location data. We demonstrate this work on a numerical example using sample data collected in the Washington D.C.-Baltimore regions that produces reasonable multimodal travel demand patterns, verified by a validation comparison to the regional household travel survey. With imputed and validated individual-level multimodal mobility, this data-driven framework will enable a series of new research topics in transportation planning, policy, health, and safety research that can never be conducted before.

The remainder of the paper will first introduce the development of the mode imputation algorithm and its related ground-truth data collection efforts in Section 2. Then, its application in a data-driven analytical framework of estimating multimodal travel demand patterns is interpreted, in the big-data application context for Baltimore-Washington D.C. regions. Results and findings with regard to the multimodal travel demand patterns are also discussed. Finally, Section 5 closes the paper with conclusions and directions for future research.

## 2. A DEEP LEARNING MODEL FOR TRAVEL MODE DETECTION

### *2.1. Benchmark: Ensemble Methods and Random Forest*
Based on the authors' previous studies on rule-based models and decision trees (Xiong and Zhang, 2013; Tang et al., 2016; Tang et al., 2017), the ensemble methods and the model of Random Forest have been tested as the benchmark for this mode detection research. An ensemble trains multiple models and combines their predictions for improved prediction accuracy (see, e.g., Hasnat and Hasan, 2018). In this paper, ensemble methods of bagging and AdaBoost have been employed and tested. A Random Forest is an ensemble of Decision Trees (Ho, 1995). For classification problems, the predictor is based on the majority voting of different trees. The training packages for ensemble and Random Forest are typically included in most existing software or platforms for artificial intelligence (e.g. *RandomForestClassifier* in Python or R).

### *2.2. A Model of Wide and Deep Learning*
In this paper, we explore the mode detection based on a wide and deep learning approach, as illustrated in Figure 1.

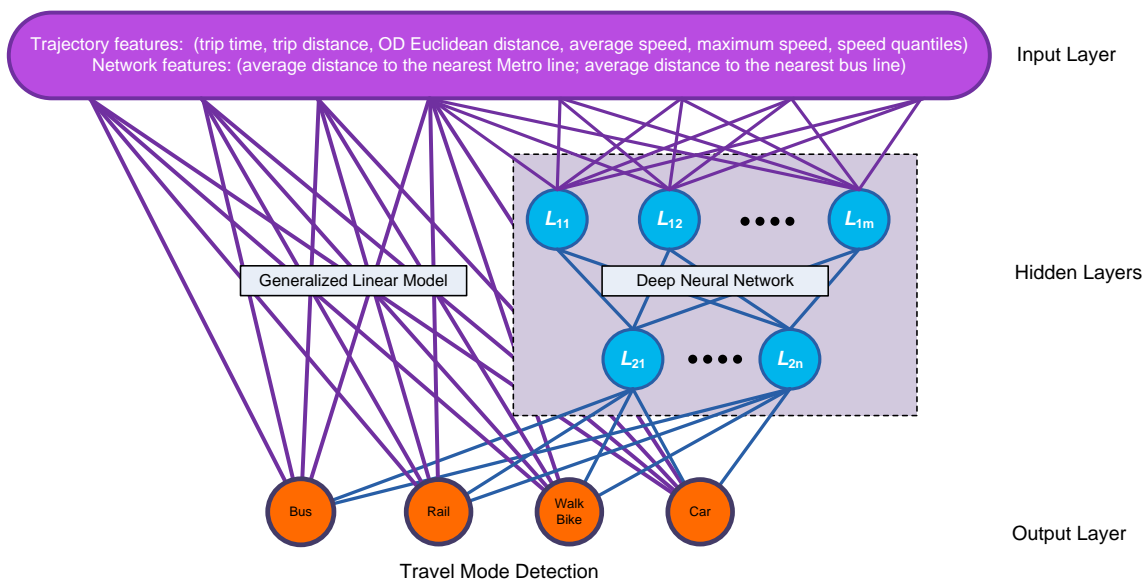

**Figure 1. The Framework of the Mode Detection Model Based on Wide and Deep Learning**

A generalized linear model and a deep neural network are jointly trained based on the passively collected data and other features. As previously discussed, this model is capable of generalizing rules and memorizing specific exceptions at the same time. It leads to a superior prediction accuracy, compared to stand-alone generalized linear models, stand-alone DNN models, and the benchmark ensemble models and Random Forest. These models are all trained and fine-tuned using the TensorFlow platform in Python.

Trajectory features (trip time, distance, OD distance, avg. speed, and maximum speed) and network features (average distances to the nearest Metro line and bus line) are used in the Wide and Deep model. These features are all continuous and normalized to the range of [0, 1]. More details of the data and variables can be found in the next subsection. Two hidden layers in the DNN are illustrated in Figure 1, with $m$ neurons and $n$ neurons, respectively. The number of layers and the number of neurons in each layer can be fine-tuned. In the empirical test of this paper, we have used three hidden layers and we have also tested the model using a different number of neurons.

Denoting $y$ as the label for travel mode, x as the vector of prediction features, beta as the vector of model parameters, and b as the unobservable heterogeneity, the wide component of the model is formulated as a generalized linear model. We specify a multinomial logit model in this case:

$$Pr(Y = y) = \frac{\exp(\boldsymbol{\beta}_y^T \mathbf{x}_y + b_y)}{\sum_i \exp(\boldsymbol{\beta}_y^T \mathbf{x}_i + b_i)} \qquad (1)$$

where $Y$ is the prediction, $\mathbf{x}_y$ is a vector of $d$ features for mode $y$, $\boldsymbol{\beta}$ is a $d$-dimensional vector of model parameters, and $b$ is the bias. Then, a three-layer deep neural network (DNN) has been specified as the deep component. For the continuous features used in the travel mode detection, including travel time, distance to Metro lines, distance to bus lines, etc., we normalize them into real-valued variables densely distributed on [0,1], with the maximum value of the variables labeled as 1. These variables are then fed into the hidden layers of the DNN to perform the following computation (see e.g. Géron, 2017 for mathematical details of the DNN formulations) in each hidden layer.

$$a^{(l+1)} = f(\gamma^{(l)} \cdot a^{(l)} + b^{(l)}) \qquad (2)$$

where $a, \gamma$, and $b$ denotes the activations, DNN parameters, and heterogeneity at the $l$-th layer respectively. $f$ denotes the activation function, which defines the output of the neuron node given an input. We use RELU (rectified linear units), $f(z) = \max(0, z)$, as the activation function. In practice, RELU function works robust and has a better computational efficiency in comparison with the other activation functions (Geron, 2017), although it is not differentiable when $z = 0$. The combination of the generalized linear model and the DNN represents a model of wide and deep learning that can be jointly trained using the weighted sum of the log-odds as the objective function. The prediction function for the wide and deep learning model is:

$$Pr(Y = y) = \sigma(\boldsymbol{\beta}_y^T \mathbf{x}_y + \gamma^{(l_f)} \cdot a^{(l_f)} + b) \qquad (3)$$

where *Pr* denotes the prediction of the joint model, $\boldsymbol{\beta}_y^T$ denotes the vector of parameters for the linear model component, and $\gamma^{(l_f)}$ denotes the finalized parameters on the final activations of the DNN component, labeled as $a^{(l_f)}$. $\sigma(\cdot)$ is the sigmoid function.

Back-propagation of the gradients is employed to jointly train the model. Gradients are defined from the mode detection to the generalized linear model and the DNN hidden layers based on the weighted sum of the log-odds from both models (Géron, 2017). A number of optimization algorithms are tested to reach the optimal level of training loss and reasonable training time at the same time, including AdaGrad (Duchi, et al., 2011), RMSProp (Tieleman and Hinton, 2012), and Adam Optimization (Kingma and Ba, 2015). RMSProp seems to yield the highest goodness of fit with acceptable computational efficiency. The models reported in this paper are trained within 20~60 seconds on a regular Macintosh machine.

AdaGrad algorithm employs adaptive learning rates with a decay factor (Duchi, et al., 2011). The rates can adapt to different gradients, which makes the algorithm suitable for high dimensional problems. However, the descent of AdaGrad can be too fast and the algorithm can get trapped in a local optimum. RMSProp and Adam algorithms address the issue by introducing an exponential decay of past gradients, so that the most recent gradient will have a higher influence on the gradient used in the current iteration. These adaptive optimization algorithms are all tested in this research to compare their performance on our mode detection dataset.

Finally, with a Random Forest model or a Wide and Deep model trained, we conduct 10-fold cross-validation to test the performance. To ensure randomness and reasonable stability of the results, we randomly sample a subset of the dataset using 10 random seeds, and then partition each subset into ten equal-sized subsamples. In each fold of the 10-fold validation, we retain one subsample as the hold-out test sample, and train the model using the remaining nine subsamples.

## *2.3. Empirical Data Collection for Travel Mode Imputation*

To train the mode imputation algorithms with ground truth observations, we have conducted a smartphone GPS survey on over 300 Washington D.C. urban travelers. A smartphone application with opt-in survey was designed to record trips for each survey subject. The survey app functions are illustrated in Figure 2:
- *GPS location tracking*: the app automatically records the user's location information. The frequency of recording is automatically adjusted based on whether the user is moving or static in order to save battery consumption. Typically, the time interval between two location records is 30 seconds when the user is moving and 10 ~ 30 minutes when the user is static depending on the battery status.
- *Opt-in trip information survey*: the app periodically pops up survey questions to record trip purposes and the travel modes for the user's recorded trips. This information is verified by a follow-up travel diary survey and then used in this research as the ground-truth travel mode dataset with labels to train the mode detection.
- *Data uploading*: for the sake of battery and cellular data usage, our app will not automatically upload data to our online database unless the device is plugged in and connected to a Wi-Fi network. Alternatively, the user could manually upload survey records by pressing the button "Press to Upload".

A total of 1009 valid trips are specified with travel mode information. The location observations of these trips are used for mode detection modeling in this study. Of these 1009 trips, 19.3% are auto trips, 15.9% are bus trips, 52.9% are metro or rail trips, and 11.9% trips are walk/bike trips. Since this survey was targeted towards urbanized areas, a higher percentage of metro and bus trips are captured. This additional bus and rail evidence helps to enhance the understanding of their characteristics and improve the model goodness-of-fit in identifying those travel modes.

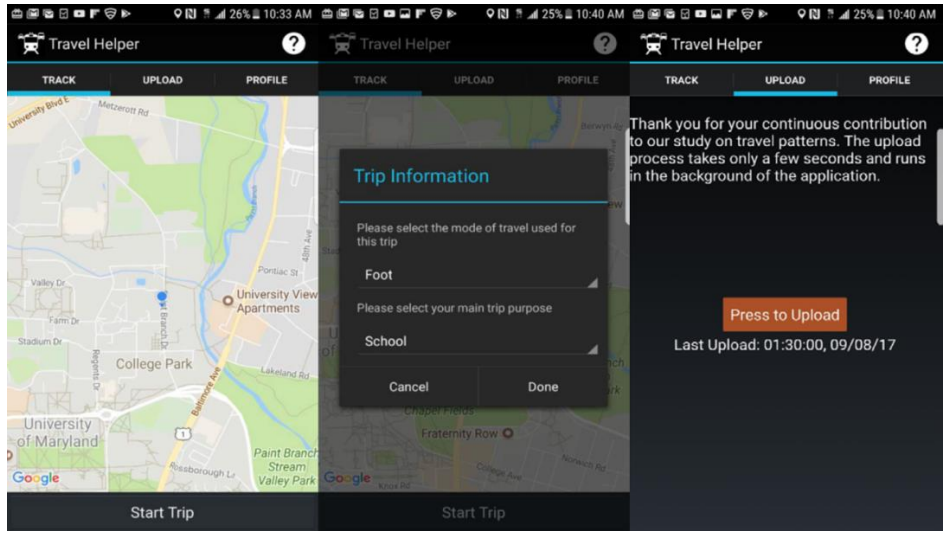

**Figure 2. The User Interface of the Smartphone GPS Data Survey App**

*Data Processing for Smartphone Survey Data and Trip End Identification*

The location point data collected by the smartphone survey in this study has information including latitude, longitude, the instantaneous speed, accuracy and timestamp. The collected raw GPS location data were filtered based on two criteria: accuracy and the average speed between two successive location points. Accuracy indicates the closeness of a measured location to the real location of the device at the time of the measurement, which is important in assessing the quality of the location data. We first filtered the data based on the accuracy and removed location data with low accuracy as an attempt to get rid of inaccurate data points. To further eliminate infeasible travel patterns, a location point is removed if the average speed between the point and its previous point is larger than 150 meters per second. This is to discard data noises (e.g. sudden "jumps" in location) and improve data quality.

To impute travel mode information, trip end information has to be extracted from a series of GPS location points. The trip end identification method in this study is similar to the approach proposed by Tang et al., (2017). A trip end is identified as the first or last location point in a stay region. In this study, a stay region is defined as the region where the user has stayed longer than a time threshold $T$, within a distance range of $D$ and under a speed limit $V$. A set of successive location points $P_l = \{p_0, p_1, \ldots, p_k\}$ are labels as a stay region if they satisfy the following constraints:

$$\Delta d_{0i} \leq D, \forall i \in P_l \tag{4}$$
$$\Delta t_{0k} \geq T \tag{5}$$
$$v_i \leq V, \forall i \in P_l \tag{6}$$

where $\Delta d_{0i}$ denotes the distance difference between the first location $p_0$ and any location $p_i$ in the location set, $\Delta t_{0k}$ is the time difference between the first and last location points, and $v_i$ represents any speed at the location $p_i$. Consequently, location $p_0$ is the trip end of the last recorded trip and $p_k$ will be the trip start of the following trip.

*The Construction of Classification Features*

As described in the modeling section, typical trajectory features are employed in our empirical testing. The following table summarizes the trajectory features. These features are selected to differentiate the modes as much as possible. For instance, the average speed can be used to distinguish walk mode from other modes. The maximum instantaneous speed further helps differentiate walk trips from auto or bus trips that encounter severe traffic congestion making their average speed close to walk trips. When the instantaneous speed is not available in the location data, we calculate the speed between any two successive points during the trip and find the maximum value. The overall data recording frequency can be used to identify metro trips as other travel modes typically do not suffer from significant GPS disruptions.

**Table 1. Mode Detection Data Description of Trajectory Features**

| Variables | Descriptions |
|---|---|
| **Trip distance** | The trip distance is computed as the sum of the distances between two successive location points in this trip |
| **Trip time** | The difference between the timestamps of the trip start and the trip end. |
| **OD Euclidean distance** | The shortest Euclidean distance between the origin and destination of the trip |
| **Average speed** | The average speed is calculated as the trip distance divided by the trip time |
| **Max. instantaneous speed** | The maximum value in the set of instantaneous speeds directly collected by the smartphone app during the trip. |
| **Speed quantiles** | The $5^{th}$, $25^{th}$, $50^{th}$, $75^{th}$, $95^{th}$ percentiles of speed are also calculated for each trip. |
| **Average data record** | The number of data points recorded during the trip divided by the trip time. |

In addition, this study also defines network features using the available metro, rail, bus, and highway networks (Figure 3). In specific, we employ the average distance to the transportation networks as the geographic features. From a location point in a trip trajectory, we first identify the nearest metro and rail line using the rail network data shown in Figure 3. The shortest Euclidean distance for each trajectory location point in that trip is calculated. These distances are then averaged to measure the average adjacency of the trip to the metro and rail systems. Similarly, we also calculate the average distance to the nearest bus line and highway network, which are deemed essential in identifying trip mode. To comprehensively assess the network effect, our rail network data was extracted from the National Transportation Atlas Database (NTAD). The General Transit

Feed Specification (GTFS) bus shapefiles have been collected from 31 regional and local agencies and bus services to construct the bus network. The highway network was built using the OpenStreetMap. The predictive power of adding these network features is assessed in the next section.

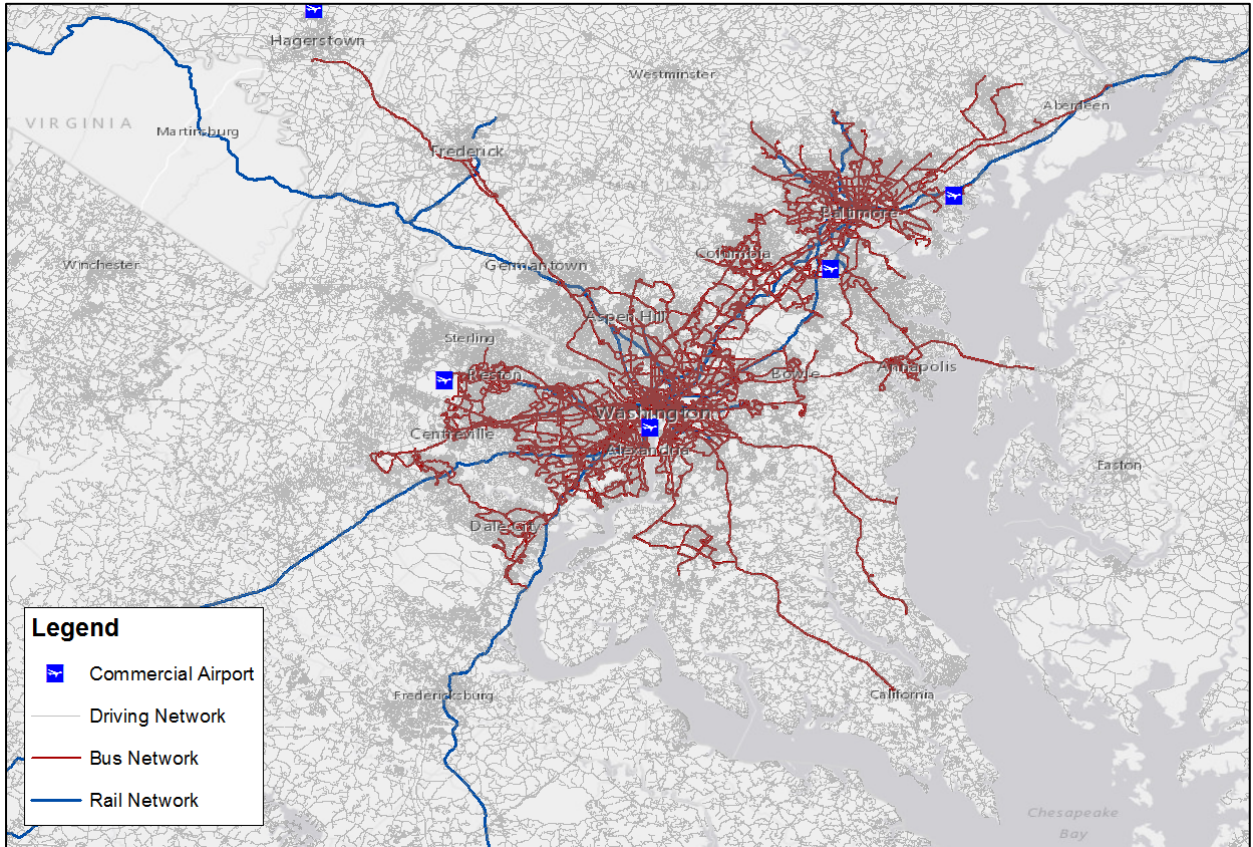

*Data sources: National Transportation Atlas Database (NTAD) for airports and rail network, the General Transit Feed Specification (GTFS) for bus network, and OpenStreetMap for highway network.*

**Figure 3. Multimodal Transportation Network of the Study Area**

## *2.4. Empirical Testing Results*

We compare the discussed models, i.e., ensemble models (AdaBoost and Bagging have been tested, Bagging is reported in this section for better performance), Random Forest, generalized linear model, deep neural network, and wide-and-deep model (various optimizers have been tested, with AdaGrad and RMSProp reported in this section), using the above-described travel mode detection dataset. We use the prediction accuracy of 10-fold cross-validation to measure the performance of the candidate models. For each round of the validation, 10 random seeds are used to ensure the stability of the validation results. Grid search and random search have been used to fine-tune the hyper-parameters in the candidate models.

Table 2 summarizes the performance measures of the models. The first finding is that the addition of multimodal network features has significantly boosted the model performance. Both the ensemble model and Random Forest have shown improved model prediction accuracy after the inclusion of network features. From the 10-fold cross-validation with 10 random seeds, a Random Forest model can get 89.6% of the travel modes in the testing data accurately detected. Also, the benchmark Random Forest model outperforms the Generalized Linear model, suggesting that rule-based generalization using features such as the maximum speed or the distance to nearby transit stations could play a significant role in travel mode detection. Compared to Random Forest, we are able to achieve a similar level of accuracy using a DNN model, given that a sufficient number of neurons is specified in the hidden layer.

The wide and deep model combines the advantages of the DNN and the Generalized Linear model, and can boost the prediction accuracy to above 95%. With 400 neuron nodes coded in the first hidden layer and a default optimizer, AdaGrad, the average prediction accuracy of the model reaches 95.7%. Compared to the benchmark model and the DNN-only model, the improvement is significant. Equivalently, the reduction of prediction errors achieved by using a joint Wide and Deep model is more than 50%. The best Wide and Deep model with RMSProp optimizer can reach 97.6% prediction accuracy. A deeper look at the confusion matrices (Table 3) will offer more insights on how these features help detect bus and Metro modes.

**Table 2. Goodness of Fit Measures for Different Travel Mode Detection Models based on Machine Learning Methods**

| Model | Total Loss | Average Loss | Average Accuracy |
|---|---|---|---|
| **Generalized Linear Model** | 26.0 | 0.299 | 0.867 |
| **Ensemble (Bagging, without network features)** | 104.4 | 1.060 | 0.755 |
| **Ensemble (Bagging, with network features)** | 84.0 | 0.860 | 0.804 |
| **Random Forest (RF, without network features)** | 52.2 | 0.600 | 0.808 |
| **Random Forest (RF, with network features)** | 17.4 | 0.193 | 0.894 |
| **Deep Neural Network (without network features)** | 15.6 | 0.175 | 0.911 |
| **Wide and Deep Model (AdaGrad Optimizer, with network features)** | 6.7 | 0.076 | 0.957 |
| **Wide and Deep Model (RMSProp Optimizer, without network features)** | 17.2 | 0.197 | 0.921 |
| **Wide and Deep Model (RMSProp Optimizer, with network features)** | 4.0 | 0.045 | 0.976 |

Compared to existing studies on mode detections, the level of accuracy that our model emits is among the top. According to the review paper by Wu et al. (2016), most studies reported an overall mode detection accuracy at around 88%~93%, while the reported highest level of accuracy is 96% (Lari, 2015). Unlike typical studies that largely draw data from auto trips (e.g. about 80% of the testing data are driving trips in Lari, 2015), our training data has relatively more balanced distribution among car, Metro, bus, and walk. Moreover, in most research papers, identifying auto trips has much higher accuracy, compared to detecting bus or Metro trips (see e.g. Lari and Golroo, 2015; Nitsche, 2014). It is not yet possible to draw a conclusion on which models perform the best without extensive examination and testing using the same datasets. However, the high prediction accuracy resulted from the Wide and Deep model, in comparison with other tested models, clearly shows its potential in handling mode detection, especially for its generalization power on the trivial differences among car, Metro, and bus trajectories.

We further compare the RF model and the proposed Wide and Deep model by examining the confusion matrices of the 10-fold cross validation (in Table 3). The sums of rows and columns may differ due to the random seeds we used. In total, we compare four models (Random Forrest (RF) and Wide and Deep (Wide-Deep), with and without the network features). From the confusion matrix, the prediction accuracy for each mode can be evaluated separately. For instance, the first row of Table 3 suggests that 195 car trips were reported in the testing dataset while 135 of them were classified correctly by the RF model (without the network features).

**Table 3. Confusion Matrix Comparison of Random Forest Model and the Wide and Deep Learning Model (RMSProp Optimizer)**

| RF without network features | | 10-Fold Cross Validation: Detected Travel Mode | | | | |
|---|---|---|---|---|---|---|
| | | Car | Metro | Bus | Walk | Recall: |
| Reported | Car | 135 | 34 | 23 | 3 | 69.2% |
| Travel | Metro | 23 | 479 | 25 | 7 | 89.7% |
| Mode | Bus | 23 | 42 | 90 | 5 | 56.3% |
| | Walk | 1 | 4 | 4 | 111 | 92.5% |
| Precision: | | 74.2% | 85.7% | 63.4% | 88.1% | 80.8% |
| **RF with network features** | | **10-Fold Cross Validation: Detected Travel Mode** | | | | |
| | | Car | Metro | Bus | Walk | Recall: |
| Reported | Car | 181 | 4 | 7 | 3 | 92.8% |
| Travel | Metro | 7 | 507 | 13 | 7 | 95.0% |
| Mode | Bus | 15 | 43 | 101 | 1 | 63.1% |
| | Walk | 0 | 5 | 2 | 113 | 94.2% |
| Precision: | | 89.2% | 90.7% | 82.1% | 91.1% | 89.4% |
| **Wide-Deep, without network features** | | **10-Fold Cross Validation: Detected Travel Mode** | | | | |
| | | Car | Metro | Bus | Walk | Recall: |
| Reported | Car | 172 | 8 | 13 | 2 | 88.2% |
| Travel | Metro | 8 | 508 | 16 | 2 | 95.1% |
| Mode | Bus | 11 | 14 | 132 | 3 | 82.5% |
| | Walk | 0 | 2 | 1 | 117 | 97.5% |
| Precision: | | 90.1% | 95.5% | 81.5% | 94.4% | 92.1% |
| **Wide-Deep, with network features** | | **10-Fold Cross Validation: Detected Travel Mode** | | | | |
| | | Car | Metro | Bus | Walk | Recall: |
| Reported | Car | 194 | 1 | 0 | 0 | 99.5% |
| Travel | Metro | 0 | 525 | 8 | 1 | 98.3% |
| Mode | Bus | 1 | 10 | 149 | 0 | 93.1% |
| | Walk | 1 | 1 | 1 | 117 | 97.5% |
| Precision: | | 99.0% | 97.8% | 94.3% | 99.2% | 97.6% |

By adding the network features, the precision and recall accuracies are significantly increased. Overall, one of the benchmark models, Random Forest with network features, did a decent job in detecting car, Metro, and walk modes. However, the precision of detecting bus mode still falls short. In fact, the accuracy of bus detection of our Random Forest model is in a similar range of the Random Forest models reported in other studies (Stenneth, et al. 2011; Lari and Golroo, 2015). Comparing the Random Forest with the Wide and Deep model, it is clear that the latter did extremely well in the detection of Metro and bus trips. Even without the network features, the Wide and Deep model can get to a similar level of accuracy of the RF model with the network features. The Wide-Deep model without network features achieves a precision accuracy of 82.5% for the bus mode, compared to 56.3% in the RF model. By adding the network features to the Wide-Deep model, the precision/recall accuracies rocket to above 93%. It is worth noting that this paper only conducted a standard grid search in combination with optimizers. By researching into the fine-tuning of the joint model, the accuracy could be further improved. This could direct our immediate next step.

## 3. THE DATA-DRIVEN ANALYTICAL FRAMEWORK

The mode detection model trained and tested in Section 2 could be used to fill an important data gap in mobile device location data, the lack of travel mode labels. With imputed travel modes, the analysis and estimation of multimodal travel demand patterns based on mobile device location data becomes feasible. To achieve such goal, a data-driven analytical framework is thus proposed as illustrated in Figure 4.

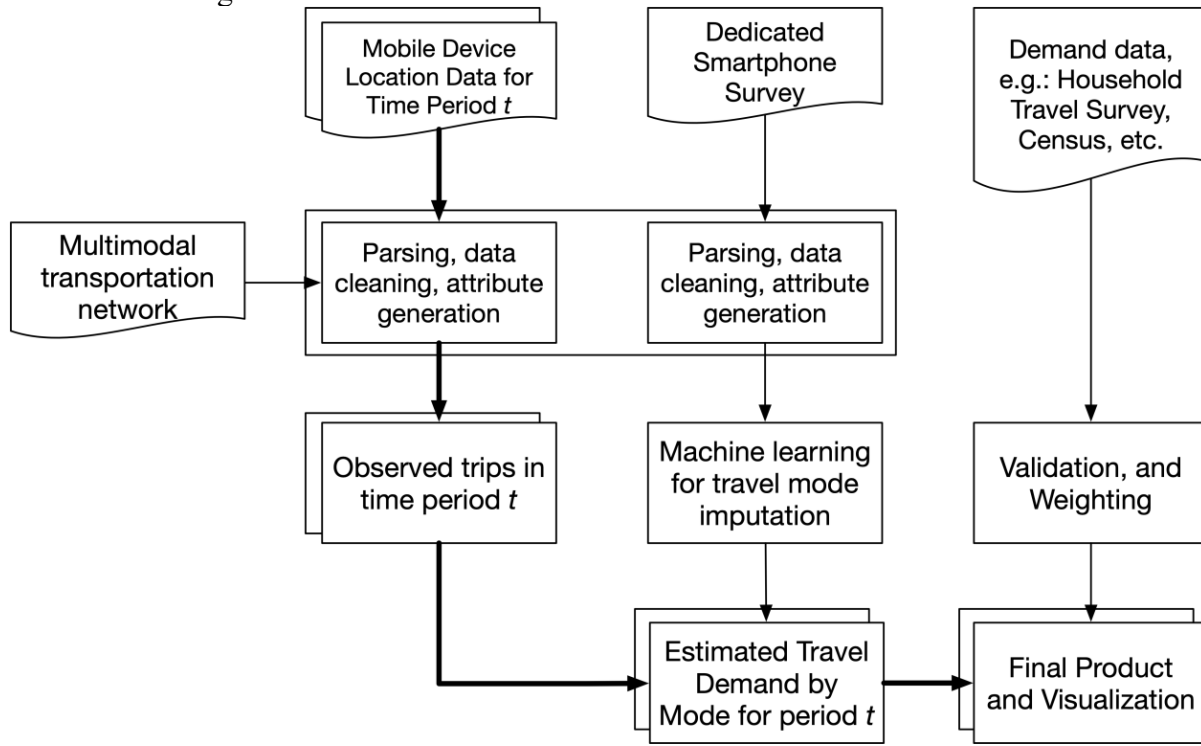

**Figure 4. The Data-Driven Analytical Framework of Estimating Multimodal Travel Demand Patterns**

The framework consists of three important pillars:
1) Central pillar: Travel mode imputation driven by the jointly trained wide-and-deep learning model. As described in Section 2, a dedicated smartphone GPS survey data provides ground-truth empirical evidence and feeds into the data parsing, preparation, and modeling. The methodology of wide-and-deep learning addresses the identification of bus, rail, and driving with satisfactory accuracy.
2) Left pillar: Big-data analytics of mobile device location data, extracting the trips and their corresponding attributes from the anonymized location points for a particular time period (July 2017, in our application), and applying the trained model on the raw location data. This will be elaborated in detail in Section 4.
3) Right pillar: Validation and weighting of the estimated travel demand. Passively-collected location data lacks ground truth to be validated against. This remains an unresolved issue for transportation planning applications using various types of mobile device data (Chen et al., 2016). The estimated travel demand by different modes can be validated using typical travel demand data sources such as household travel surveys and traffic counts.

These pillars are completely data-driven, taking advantage of the rich mobile device location data. To the authors' best knowledge, this framework is one of the first in the transportation domain that produces validated multimodal travel demand estimates. The bold arrows in Figure 4 indicate the components that are dynamically injected based on the data derived specifically for a time period $t$. This allows aggregate/disaggregate travel demand analysis in a customizable time horizon.

## 4. APPLICATION CONTEXT: ESTIMATING THE MULTIMODAL TRAVEL DEMAND

### 4.1. The mobile device location data for Baltimore-D.C. Metropolitan Regions

After developing the mode imputation algorithm, we now present an application of the developed algorithm to the real-world raw mobile device location data. We focus on Washington D.C. and Baltimore Metropolitan Regions, covering 10 counties and over 5.5 million population. The mobile device location data is generated by location-based service (LBS) that smartphone apps use to determine users' location. Each data record contains device ID, latitude, longitude, and a timestamp. The data is continuous as long as the app has access to location service and it may work in the background. We process one month of the anonymized mobile device location data provided by AirSage Inc. The location dataset includes more than 45 million data records produced by nearly 350,000 mobile phones in July, 2017.

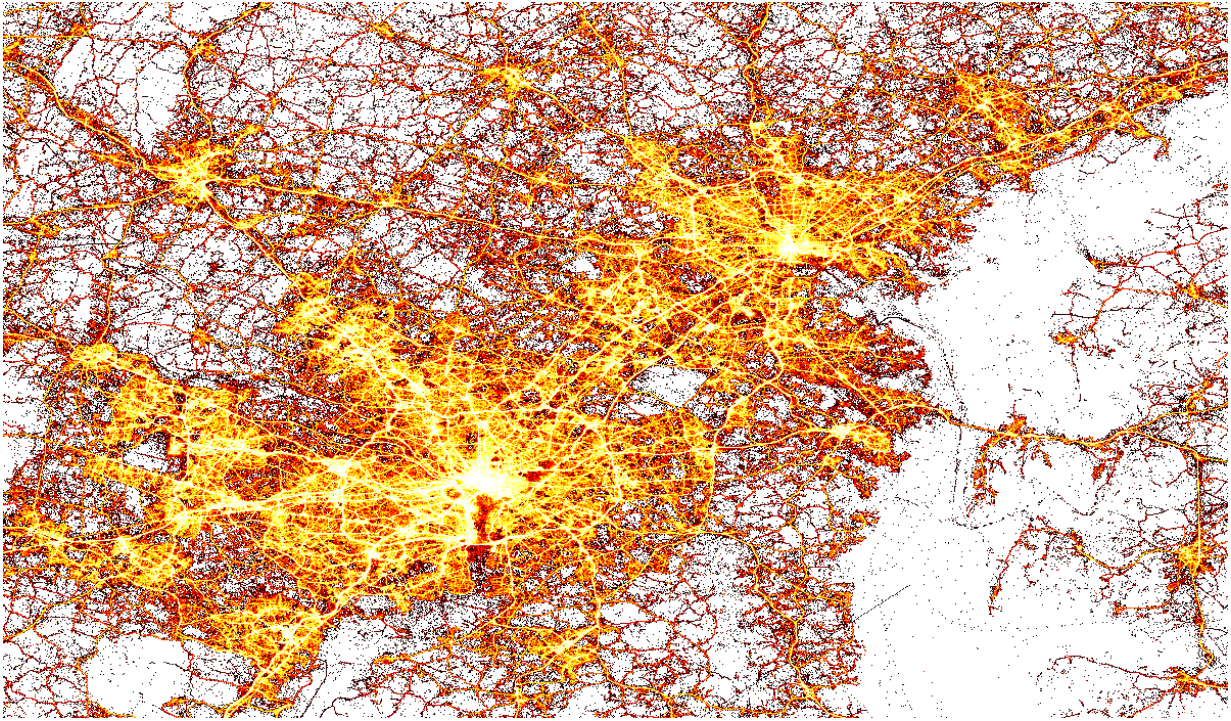

**Figure 5. Density Map of Anonymized Data Records of the Mobile Device Location Data (July 2017 Data for Washington D.C. and Baltimore Metropolitan Regions Used)**

*4.2. Trip Identification for the Anonymized Mobile Device Location Data*

As previously mentioned, the mode imputation algorithm trained on the collected labeled smartphone GPS survey data needs to be applied to raw mobile device location data. The raw data is not labeled, even the trips are not identified. The first step, before trip attribute extraction and mode imputation is to identify trips from raw data. Figure 6 shows the trip identification algorithm used in this paper. This algorithm is a simple efficient algorithm based on the evaluation of consecutive observations by a device. Before applying the algorithm, all devices with a single observation are removed from the dataset as a part of data cleaning. For each device, the algorithm assigns a trip ID to the first observation of the device. Set this first point as the current point. Then the algorithm checks the next point of the device to see if it is in the same trip with the current point, based on the attributes *distance-from*, *speed-from*, and *time-from* of the current point. *Distance-from* is the distance from the current point to its next point if there is a next point; otherwise, it is set as 0. *Speed-from* and *time-from* are defined in a similar fashion. These attributes are compared with their corresponding thresholds to evaluate if the next point is in the same trip with the current point. The thresholds can be seen as hyper parameters which need to be set before the application of the algorithm. These parameters can be selected through prior knowledge or selected based on the cross-validation using comparison with aggregate-level ground-truth trip rates and trip length distribution. These parameters can be calibrated to better match ground truth data.

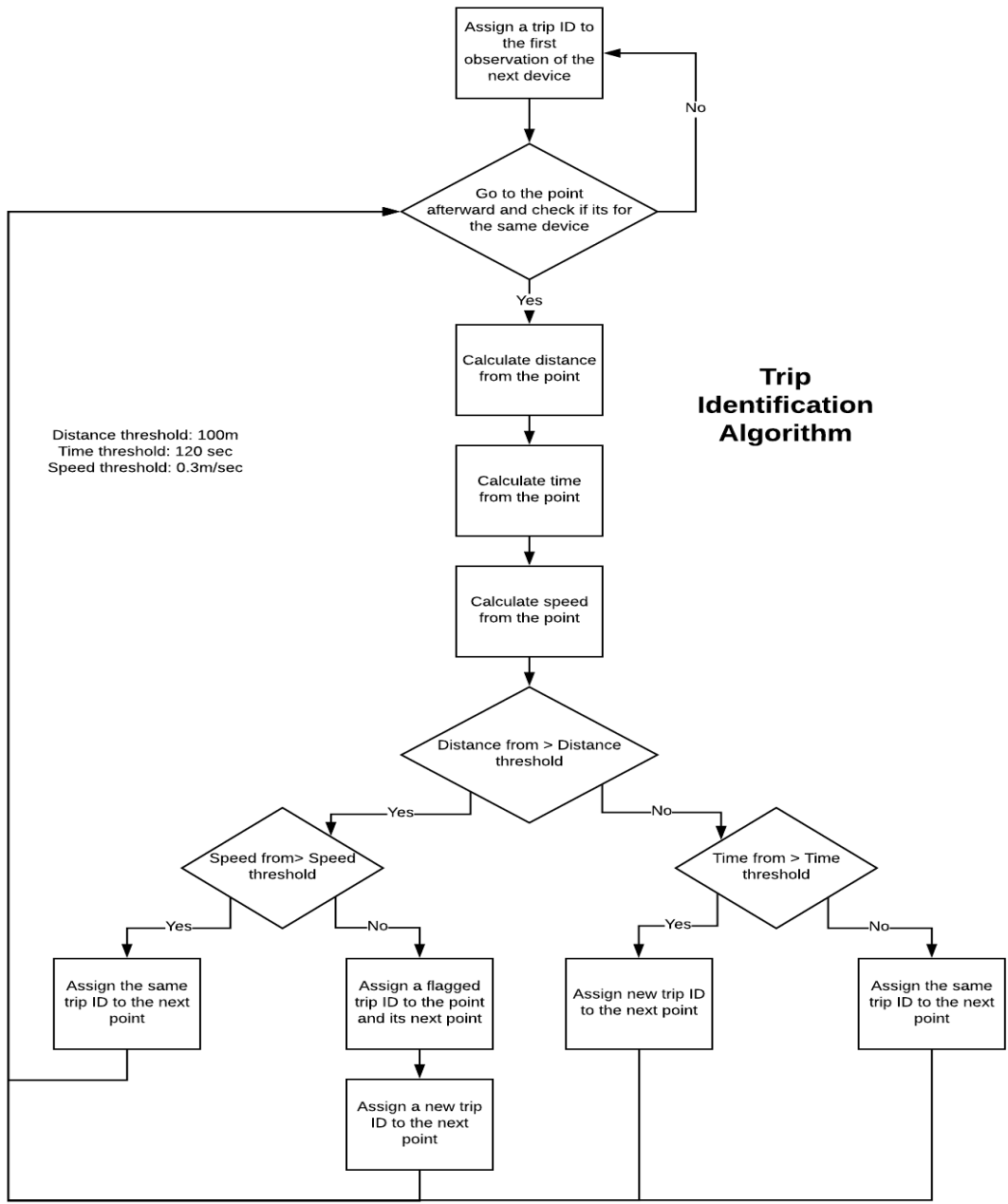

**Figure 6. Trip Identification for the Mobile Device Location Data**

If the algorithm identifies that the next point is in the same trip, the same trip ID is assigned to the next point and this point becomes the current point; otherwise, a new trip ID will be assigned to the next point and this point becomes the current point. The same evaluation will now be made for the new current point to evaluate if its next point is recorded on the same trip with it. Once all device points are evaluated, trip IDs with a single location point are removed, as they, in fact, do not belong to any trip.

We apply the trip identification algorithm to the raw locations data and get around 5.8 million imputed trips for the D.C.–Baltimore regions. The trip volume is visualized by county in Figure 7. The volume distributions by trip origins and destinations are very similar, with the Montgomery County in Maryland having the highest volume—more than 600,000—of trips.

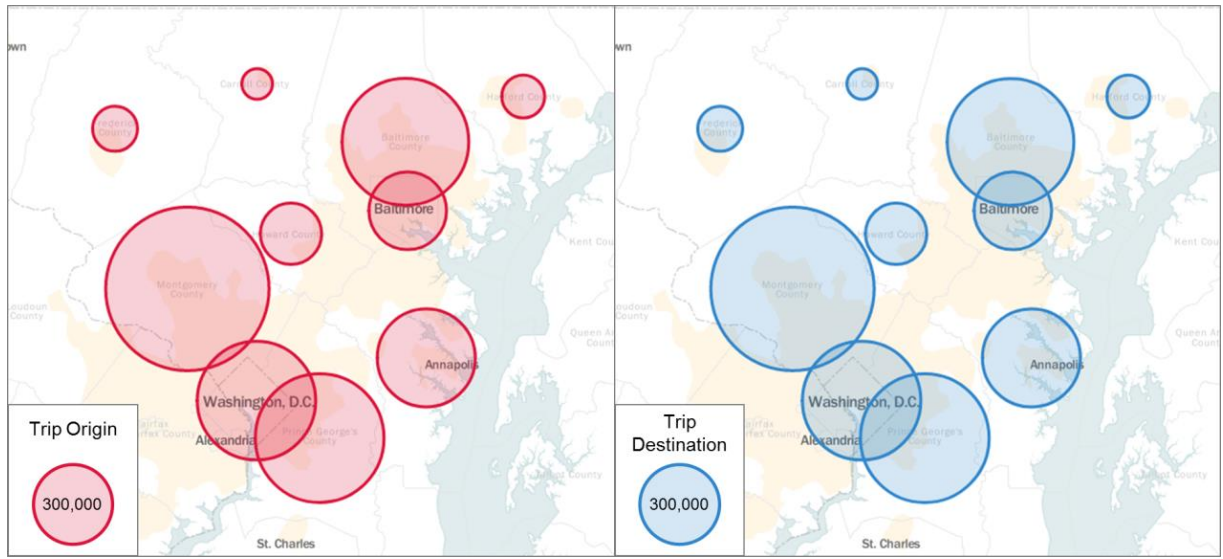

**Figure 7. The Volumes of Trip Records by County**

*4.3. Mode Imputation Results*

Applying the process of parsing, attribute generation, and mode imputation to these mobile device trips, we have generated the multimodal travel demand patterns of the study area for July 2017. Visualized in Figure 8, these trips are labeled with travel modes: highway (76.0% of the total number of trips observed, colored in blue), rail (5.9%, colored in green), bus (7.7%, colored in red), and walk/bike (10.4% colored in yellow).

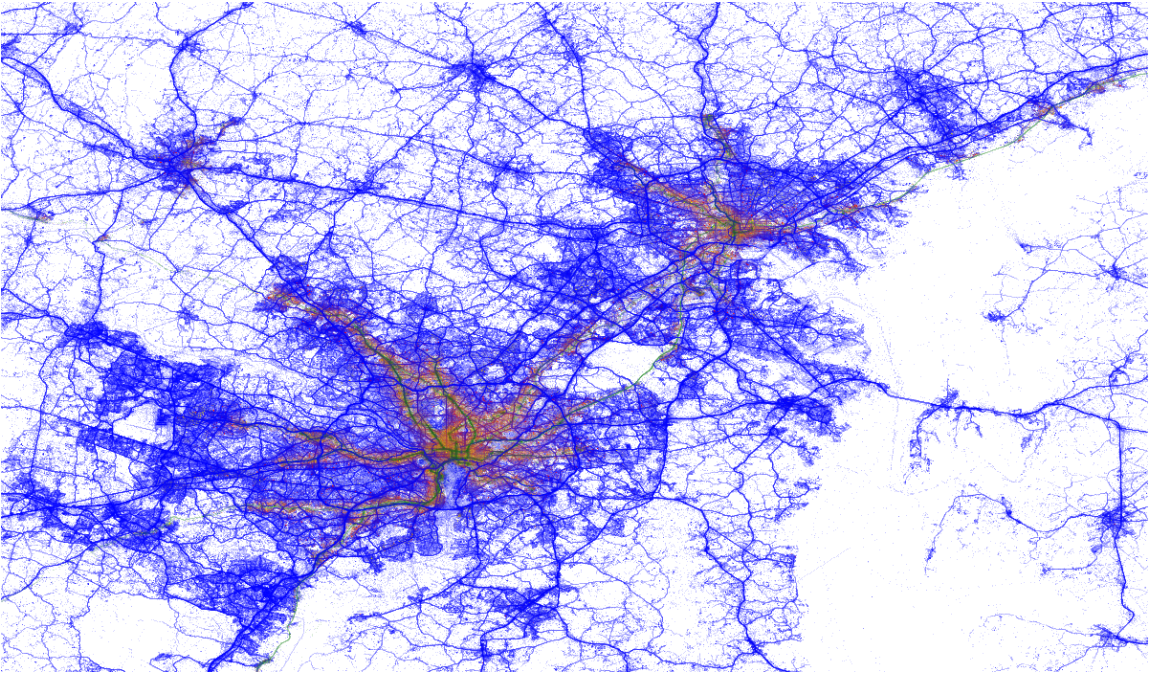

**Figure 8. Multimodal Travel Demand Patterns in the Washington D.C. and Baltimore Regions (Blue: Highway Trips; Green: Metro and Rail Trips; Red: Bus Trips; Yellow: Walk/Bike Trips. July 2017 Data Demonstrated).**

*4.4. Validation*

In our validation process, we tried to leverage an independent dataset to examine the performance of the presented framework. Considering the study area of the application, the 2007-2008 Transportation Planning Board-Baltimore Metropolitan Council (TPB-BMC) Household Travel Survey (HTS) is the latest and most comprehensive data source available. The survey was conducted in both Baltimore and Washington regions in 2007/2008 including nearly 15,000 households. The Survey data are organized in four relational databases described as household file, person file, trip file, and vehicle file. As the first step of the validation, the aggregate mode share results from the imputed trips were compared against the weighted mode share results (weighted using population data) of the survey (Table 4). As it can be seen, there is a good match between the imputed trips and TPB-BMC survey for all the modes. An underestimation exists for the highway mode while other mode shares have been slightly overestimated. These discrepancies can be attributed to several facts. The first possible reason could be the temporal difference between the two datasets as the imputed trips are for July 2017 and the survey represents the travel pattern in 2007/2008. The other reason to argue is that transit services have been improved during the past 10 years especially in Washington D.C. and Baltimore city area and these improvements might have led to an increase in transit modes share. Furthermore, it should be also noted that the imputed trips only represent the travel patterns for the month of July. The good weather during this month might also lead to an increase in the number of non-motorized trips comparing to the survey results which represents an average day.

**Table 4. Trip Mode Share Validation against TPB-BMC HTS Survey (Weighted using Population Data)**

| Mode Share | Imputed Trips from Big Data | Weighted Trips from TPB-BMC HTS Survey |
|---|---|---|
| Highway | 76.00% | 80.5% |
| Rail | 5.90% | 4.1% |
| Bus | 7.70% | 5.7% |
| Non-Motorized | 10.40% | 9.7% |
| Total Number of Trips | 5,816,624 | 27,890,344 |

In addition to the aggregate level mode share comparison between the imputed mode results and the TPB-BMC HTS survey, the survey was used to evaluate the consistency of the trip characteristics for each mode. In order to delve more into the validation of both trip identification and trip mode algorithms, trip length distribution and travel time distribution were investigated (Figure 9 and 10). In terms of travel time distribution, highway and non-motorized travel time distribution follow the same pattern in both imputed trips and survey data. However, variations can be seen in travel time pattern for rail and bus modes. These discrepancies can be attributed to the fact that there are limitations in mobile device data. In particular, signal loss and urban canyon effect might be the main reasons for these mismatches since metro and rail trips are significantly more prone to these effects. In our framework, if signal loss happens during the trip, the trip time will be estimated using the starting point and the last observation before the signal loss occurs which makes the transit trips tend to have shorter travel time in comparison to the survey.

As the next step of our validation, trip length distributions were examined for each mode (Figure 10). In terms of trip length distribution, imputed trips' patterns have great conformity with travel survey patterns except for the rail mode. For the rail mode case, again our framework observed more short distance trips in comparison to what survey denotes. This discrepancy is expected as discussed for the trip time distribution due to the signal loss effect which prevents us to capture the entire path of the rail trips in some cases.

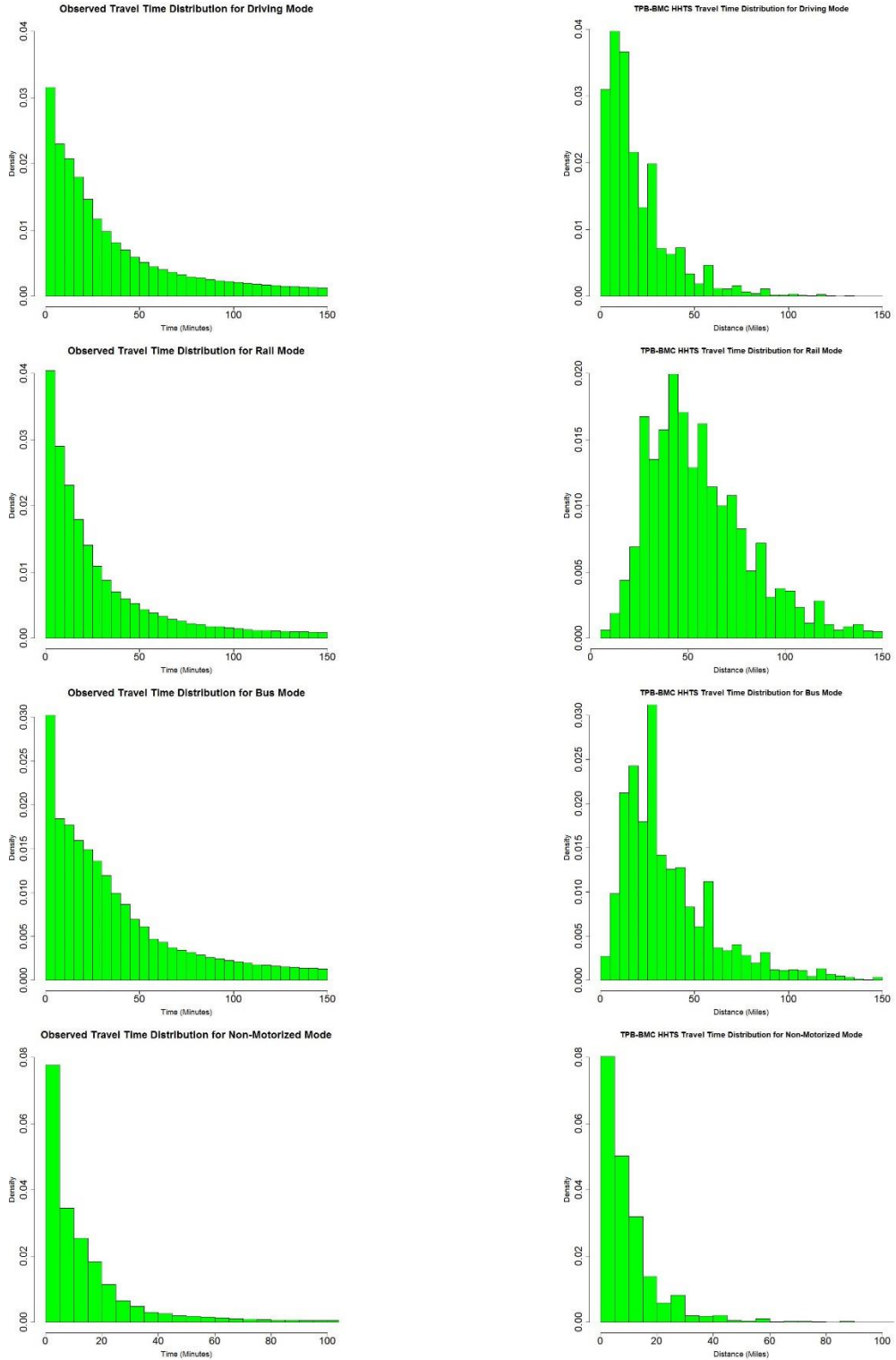

**Figure 9. Travel Time Distribution Comparison between Imputed Trips and TPB-BMC HTS Travel Survey for Each Mode (The Left Column Is for Imputed Trips using Our Presented Framework and The Right Column Shows The Results from The Survey)**

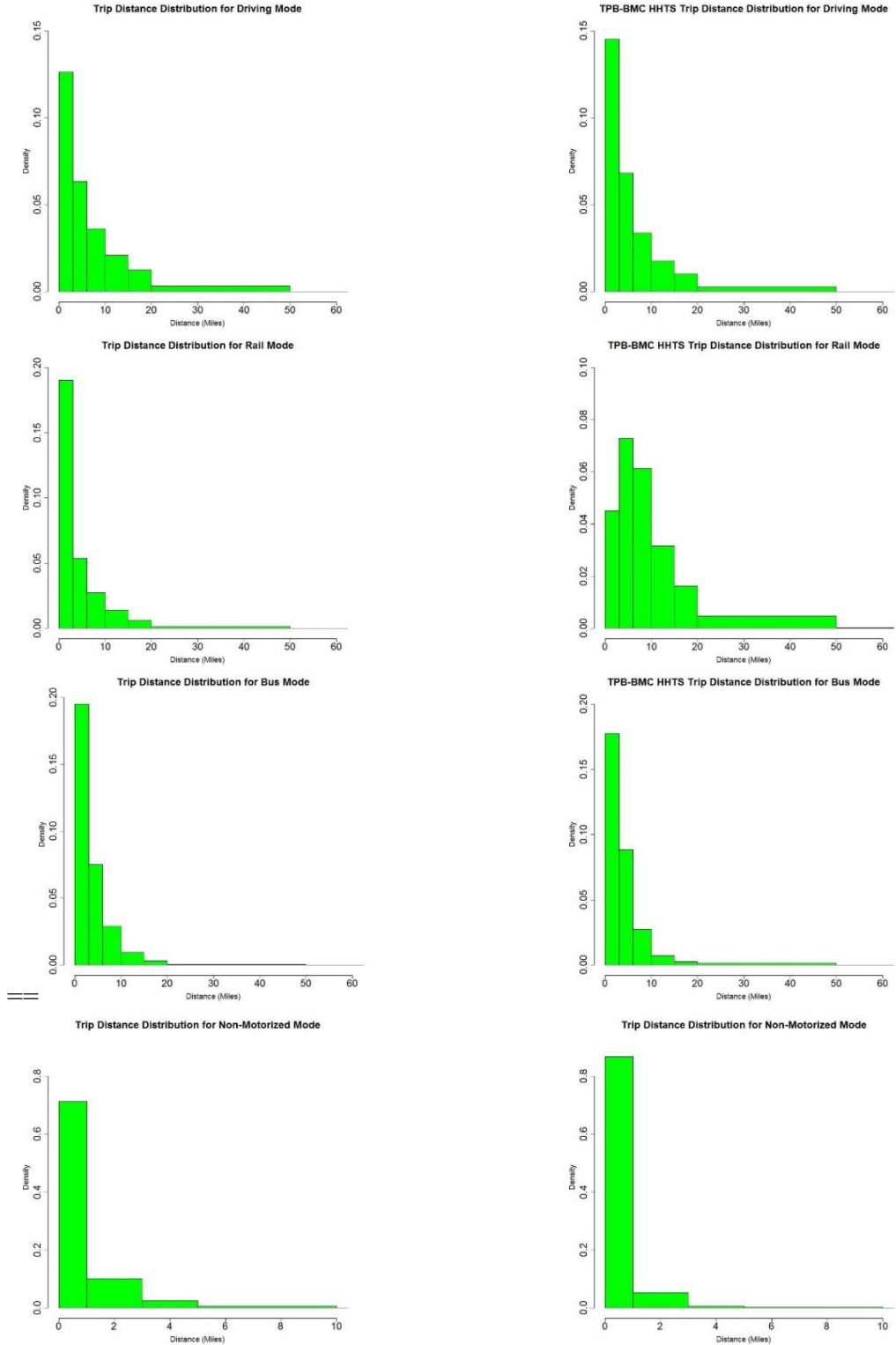

**Figure 10. Trip Distance Distribution Comparison between Imputed Trips and TPB-BMC HTS Travel Survey for Each Mode (The Left Column Is for Imputed Trips using Our Presented Framework and The Right Column Shows The Results from The Survey)**

*4.5. Additional Findings and Visualizations*

We have developed visualizations of the multimodal travel patterns estimated for the two urban areas, Washington and Baltimore, as depicted in Figure 11 and 12. While highway travels are still dominating, especially in the suburban areas, we see a significant amount of mobility using other transportation modes in the urban areas of the two cities. Washington D.C. has a highly utilized Metro system. Dense metro/rail trips (marked in green) are seen in the CBD area, as well as remote suburban centers including Bethesda and Silver Spring in Maryland (the two clusters near the top boundary of Figure 11) Arlington and Crystal City in Northern Virginia (the two clusters at the bottom-left corner of Figure 12). Bus and walk/bike mobility are distributed in the urban area as well as the buffering area of the Metro system, serving the first and last-mile travel needs. In addition, more estimated bus and walk/bike trips are observed in the eastern part of the Washington D.C. and Prince George's County in Maryland.

At Baltimore (Figure 12), we see more concentrated walking and biking at the core CBD areas and several major land use and public service areas, including the Camden Yards and the Johns Hopkins Hospital (the biggest yellow cluster to the east of CBD Baltimore). Rail and metro trips are also concentrated in Metro SubwayLink and Light RailLink segments that serve the downtown Baltimore area. Compared to Washington D.C., fewer rail/metro travels are estimated for the remote service areas, i.e. the Northwest and Southwest Baltimore in Figure 12. On the spatial distribution of multimodal trips, we estimate that West Baltimore has more transit and walk/bike trips, compared to its northern and eastern counterparts. All these model results match the expectations in these neighborhoods.

In this visualization demonstration, we have used the data for July 2017. Seasonal travel patterns should also be considered in the results. For instance, we estimate that the mode share for walk and bike trips is 10.4%, which is slightly higher than the observed mode share from the D.C.-Baltimore household travel survey. This could be attributed to the fact that more walking and biking occur in summer time when an extensive amount of outdoor activities and tourist visits are expected in the region. Another example is the reduced amount of trips observed near university campuses, e.g. the University of Maryland (located at the top-right corner of Figure 11) and Towson University (located in northern Baltimore). This, as well as other interesting dynamics in multimodal travel demand patterns, can be further examined and verified when this framework is extended spatially and temporally.

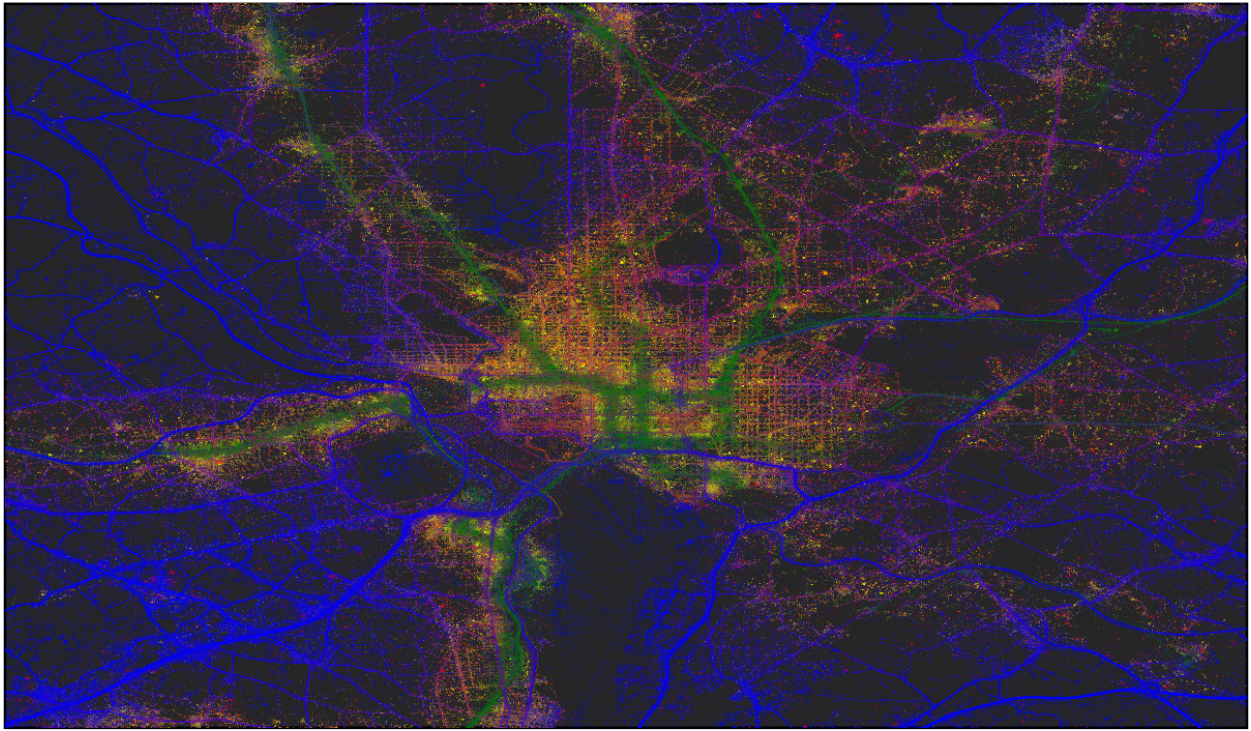

**Figure 11. Multimodal Travel Demand in Washington D.C. (Blue: Highway Trips; Green: Rail/Metro; Red: Bus; Yellow: Walk/Bike; July 2017 Data Used.)**

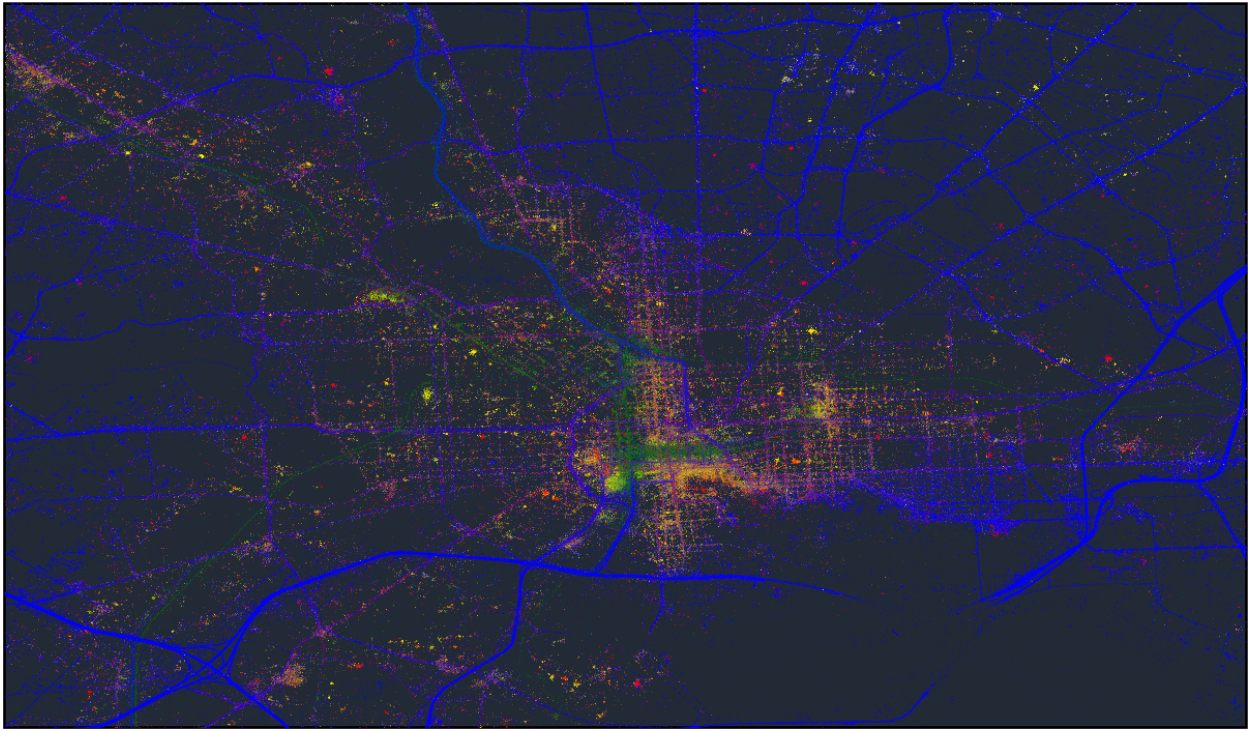

**Figure 12. Multimodal Travel Demand in Baltimore (Blue: Highway Trips; Green: Rail/Metro; Red: Bus; Yellow: Walk/Bike; July 2017 Data Used.)**

# 5. CONCLUSION AND DISCUSSIONS

This paper develops a data-driven analytical framework to analyze the mobile device location data and estimate multimodal travel demand. Travel modes for the observed trips in the big data are imputed using an empirically trained travel mode detection model. To train this model, a GPS-based travel survey has been conducted in the Washington D.C. and Baltimore regions. The framework has been applied to a mobile device location dataset that contains more than 5 million trips observed in July 2017 in the D.C.-Baltimore regions. We visualized the outcomes for the two major cities in the area and validated the results against the observations in the regional household travel survey. The travel demand products are communicated via a web-based interactive visualization platform to agencies, researchers, and the public.

The major contribution of this study is two folds. Methodologically, it is one of the first studies that developed a jointly-trained wide and deep learning approach for detecting travel mode using mobile device location data. The method has been empirically tested on a sample of over 1,000 observed trips with mode labels. Being "wide" and "deep" at the same time, this proposed model combines the advantages of both types of models and greatly improves the model goodness-of-fit. To further improve the accuracy, the paper adopted multimodal network measurements and assembled all available GTFS bus data from 31 regional and local agencies and bus service providers. These features were found extremely useful to detect bus trips and Metro trips that contain underground portions and/or have signal quality issues. These collectively contribute to a model with high detection accuracy at the individual trip level. It boosts the prediction accuracy of individual trip mode from around 90% for the RF models to above 97%, while the detection of bus mode has above 93% recall and precision accuracy.

To advance both the state-of-the-art and the state-of-the-practice in using mobile device location data for estimating multimodal travel demand, we then developed a data-driven analytical framework and successfully applied the framework on a sample mobile device data collected for July 2017 in the Washington D.C.-Baltimore regions. With the wide-and-deep learning for mode imputation plugged in, the framework estimated reasonable multimodal travel demand patterns, verified by a validation comparison to the regional household travel survey. It represents a first attempt to fill a critical gap of lacking modal information in the mobile device location data. Moreover, this demonstrated data product has several advantages. It has more frequent and accurate location information compared to cellphone call records. Volumes on different modal infrastructures can thus be inferred with confidence. This analytical framework will enable extensive research in urban computing, transportation planning and policy, human behavior dynamics, and more. In our application context, it is already demonstrated that travelers in certain areas of the Baltimore-Washington D.C. region rely more heavily on public transit and walk/bike. The model can be applied to assess the equity issue and evaluate the actual mobility/accessibility of citizens to jobs, food, and health care services. It can also boost research in travel behavior and public health. The model can accurately detect the duration and intensity of active travel, such as walking, biking, and running, which are important statistical indicators of overall health level of an individual.

Being exploratory research, this study is limited in several aspects. The data pre-processing of the mobile device location data can be further explored. As we noted in the validation, the big data contains fragmentations where it could be the case that only a small number of records were observed for a particular trip. This is more frequently seen in metro and rail trajectories due to the loss or sudden jumps of GPS signal. To address this limitation, one can explore the reconstruction of trip trajectories based on fragmented data (e.g., Wang and Chen 2018) or try to infer the trip origin-destination based on a-priori information observed for the same mobile device in the big dataset (e.g., Bachir et al., 2019). Moreover, to turn the estimated multimodal travel demand into useful origin-destination matrices, appropriate weighting method should be developed to supplement the existing transportation planning process.

## ACKNOWLEDGEMENT

The research is partially financially supported by a USDOT Federal Highway Administration (FHWA) Exploratory Advanced Research (EAR) project entitled "Data Analytics and Modeling Methods for Tracking and Predicting Origin-Destination Travel Trends based on Mobile Device Data". The opinions in this paper do not necessarily reflect the official views of USDOT or FHWA. They assume no liability for the content or use of this paper. The authors are solely responsible for all statements in this paper. We thank Liang Tang and Di Yang for their contribution in the GPS-survey data collection and data processing. We thank Josef Cigoi, Jonathan Apple, and Vijay Sivaraman, and Sashi Gurram at Airsage Inc. for their assistance in mobile device location data query and data processing.

## REFERENCES:

Bachir, D., Khodabandelou, G., Gauthier, V., Yacoubi, M., Puchinger, J. (2019). Inferring dynamic origin-destination flows by transport mode using mobile phone data. Transportation Research Part C. 101, 254-275.

Belik, V., Geisel, T., Brockmann, D. 2011. Natural human mobility patterns and spatial spread of infectious diseases. Phys. Rev. X 1(1): 011001.

Byon, Y.; Liang, S. (2014). Real-time transportation mode detection using smartphones and artificial neural networks: Performance comparisons between smartphones and conventional global positioning system sensors. J. Intell. Transp. Syst. 2014, 18, 264–272.

Calabrese, F., Diao, M., Di Lorenzo, G., Ferreira Jr., J., Ratti, C., 2013. Understanding individual mobility patterns from urban sensing data: a mobile phone trace example. Transportation Research Part C. 26. 301-313.

Chen, C., Ma, J., Susilo, Y., Liu, Y., & Wang, M. (2016). The promises of big data and small data for travel behavior (aka human mobility) analysis. Transportation Research Part C: Emerging Technologies, 68, 285-299.

Cheng, H. T., Koc, L., Harmsen, J., Shaked, T., Chandra, T., Aradhye, H., ... & Anil, R. (2016). Wide & deep learning for recommender systems. In Proceedings of the 1st Workshop on Deep Learning for Recommender Systems (pp. 7-10). ACM.

Dabiri, S., Heaslip, K. (2018). Inferring transportation modes from GPS trajectories using a convolutional neural network. Transportation Research Part C. 86, 360-371.

Daganzo, C.F. (1980). Optimal sampling strategies for statistical models with discrete dependent variables. Transportation Science. 14(4). 324-345.

De Montjoye, Y.A., Hidalgo C.A., Verleysen, M., Blondel, V.D. (2013). Unique in the crowd: the privacy bounds of human mobility. Scientific Report. 3.

Duchi, J., Hazan, E., Singer, Y. (2011). Adaptive subgradient methods for online learning and stochastic optimization. Journal of Machine Learning Research, 12:2121-2159.

Géron, A. (2017). Hands-on machine learning with Scikit-Learn and TensorFlow: concepts, tools, and techniques to build intelligent systems. O'Reilly Media, Inc. Sebastopol, CA.

Gonzalez, M.C., Hidalgo, C.A., Barabasi, A.L. 2008. Understanding individual human mobility patterns. Nature 453(7196), 779-782.

Gonzalez, P.A.; Weinstein, J.S.; Barbeau, S.J.; Labrador, M.A.; Winters, P.L.; Georggi, N.L.; Perez, R. (2010) Automating mode detection for travel behaviour analysis by using global positioning systems-enabled mobile phones and neural networks. IET Intell. Transp. Syst. 2010, 4, 37–49.

Hasnat, M. M., & Hasan, S. (2018). Identifying tourists and analyzing spatial patterns of their destinations from location-based social media data. Transportation Research Part C: Emerging Technologies, 96, 38-54.

Ho, T. K. (1995). Random decision forests. Proceedings of the Third International Conference on Document Analysis and Recognition, 1, 278-282). IEEE.

Huang H., Cheng, Y., Weibel, R. (2019). Transport mode detection based on mobile phone network data: A systematic review. Transportation Research Part C. 101: 297-312.

Iqbal, M. S., Choudhury, C. F., Wang, P., & González, M. C. (2014). Development of origin–destination matrices using mobile phone call data. Transportation Research Part C: Emerging Technologies, 40, 63-74.

Kingma, D., Ba, J. (2015). Adam: A Method for Stochastic Optimization. ICLR 2015 Conference Paper. https://arxiv.org/pdf/1412.6980v9.pdf

Lari, Z.A. and Golroo, A. (2015). Automated Transportation Mode Detection Using Smart Phone Applications via Machine Learning: Case Study Mega City of Tehran. In Proceedings of the Transportation Research Board 94th Annual Meeting, Washington, D.C., 11~15 January 2015.

Liao, L., Patterson, D. J., Fox, D., & Kautz, H. (2007). Learning and inferring transportation routines. Artificial Intelligence, 171(5-6), 311-331.

Louppe, G. Understanding Random Forests: From Theory to Practice. July, 2014. Ph.D. Thesis, University of Liege. https://doi.org/10.13140/2.1.1570.5928

Lu, X., Bengtsson, L., Holme, P. 2012. Predictability of population displacement after the 2010 Haiti earthquake. Proc. Nat. Acad. Sci. 109(29), 11576-11581.

Nitsche, P.; Widhalm, P.; Breuss, S.; Brändle, N.; Maurer, P. Supporting large-scale travel surveys with smartphones—A practical approach. Transp. Res. Part C Emerg. Technol. 2014, 43, 212–221

Pendyala, R., Kitamura, R., Kikuchi, A. et al., (2005). Florida Activity Mobility Simulator: Overview and Preliminary Validation Results. Transportation Research Records. https://doi.org/10.1177/0361198105192100114

Richardson, A.J., Ampt, E.S., Meyburg, A.H., (1995). Survey Methods for Transport Planning. Eucalyptus Press. Melbourne.

Schneider, C.M., Belik, V., Couronne T., Smoreda, Z., Gonzalez, M.C. 2013. Unravelling daily human mobility motifs. Journal of Royal Society Interface. 10(84): 20130246.

Stenneth, L.; Wolfson, O.; Yu, P.S.; Xu, B. (2011). Transportation Mode Detection Using Mobile Phones and GIS Information. In Proceedings of the 19th ACM SIGSPATIAL

International Conference on Advances in Geographic Information Systems, Chicago, IL, USA, 1–4 November 2011.

Stopher, P.R., Greaves, S.P. (2007). Household travel surveys: where are we going? Transportation Research Part A. 41(5), 367-381.

Tang, L., Pan, Y., and Zhang, L. (2018). Trip purpose imputation based on long-term GPS data. In proceedings of the 2018 Transportation Research Board Annual Meeting. Jan 8-11, 2018, Washington D.C.

Tang, L., Xiong, C., and Zhang, L. (2015). Decision tree method for modeling travel mode switching in a dynamic behavioral process. Transportation Planning and Technology. 38(8), 833-850.

Tieleman, T., and Hinton, G. (2012). Lecture 6.5-RMSProp: Divide the gradient by a running average of its recent magnitude. COURSERA: Neural networks for machine learning, 4(2), 26-31.

Toole, J., Colak, S., Sturt, B., Alexander, L., Evsukoff, A., Gonzalez, M.C. (2015). The path most traveled: travel demand estimation using big data sources. Transportation Research Part C. 2015. https://doi.org/10.1016/j.trc.2015.04.022

Wang, F., Chen, C. (2018). On data processing required to derive mobility patterns from passively-generated mobile phone data. Transportation Research Part C. 87, 58-74.

Wesolowski, A., Eagle, N., Tatem, A.J., Smith D.L. Noor A.M. Snow, R.W., Buckee, C.O., (2012). Quantifying the impact of human mobility on malaria. Science. 338 (6104), 267-270.

Witayangkurn A, Horanont T, Ono N, Sekimoto Y, Shibasaki R. Trip reconstruction and transportation mode extraction on low data rate GPS data from mobile phone. InProceedings of the international conference on computers in urban planning and urban management (CUPUM 2013) 2013 Jul (pp. 1-19).

Wu, L., Yang, B., Jing, P. (2016). Travel mode detection based on GPS raw data collected by Smartphones: A systematic review of the existing methodologies. *Information*. 7 (67). doi:10.3390/info7040067

Xiao, G.; Juan, Z.; Gao, J. (2015) Travel Mode Detection Based on Neural Networks and Particle Swarm Optimization. Information 2015, 6, 522–535.

Xiong, C. and Zhang, L. (2013). A descriptive Bayesian approach to modeling and calibrating drivers' en-route diversion behavior. IEEE Transactions on Intelligent Transportation Systems. 14(4). 1817-1824.